\documentclass[10pt,twocolumn,letterpaper]{article}

\usepackage{cvpr}
\usepackage{times}
\usepackage{epsfig}
\usepackage{graphicx}
\usepackage{amsmath}
\usepackage{amssymb}

\usepackage{epsfig}
\usepackage{subfigure}
\usepackage{graphicx}
\usepackage{algorithm}
\usepackage{algorithmic}
\usepackage{amssymb}
\usepackage{mathrsfs} % For LaTeX2e
\usepackage{amsmath,bm}
\usepackage{array,multirow}
\usepackage{mdwlist}
\usepackage{paralist}
\usepackage{url}
\usepackage{color}
\usepackage{caption}
\usepackage{setspace}
\usepackage{textcomp}
\usepackage{bbding}
\usepackage{booktabs}
\usepackage{threeparttable}

\usepackage[pagebackref=true,breaklinks=true,letterpaper=true,colorlinks,bookmarks=false]{hyperref}

\cvprfinalcopy % *** Uncomment this line for the final submission

\def\cvprPaperID{5} % *** Enter the CVPR Paper ID here

\ifcvprfinal\fi
\begin{document}
%\hyphenpenalty=400

%%%%%%%%% TITLE
\title{Single-Shot Refinement Neural Network for Object Detection}

\author{Shifeng Zhang$^{1,2}$, Longyin Wen$^3$, Xiao Bian$^3$, Zhen Lei$^{1,2}$, Stan Z. Li$^{1,2}$\\
$^1$ CBSR \& NLPR, Institute of Automation, Chinese Academy of Sciences, Beijing, China. \\
$^2$ University of Chinese Academy of Sciences, Beijing, China. \\
$^3$ GE Global Research, Niskayuna, NY. \\
{\tt\small \{shifeng.zhang,zlei,szli\}@nlpr.ia.ac.cn}, {\tt \small \{longyin.wen,xiao.bian\}@ge.com}
}

\maketitle
%\thispagestyle{empty}

%%%%%%%%% ABSTRACT
\begin{abstract}

For object detection, the two-stage approach (\eg, Faster R-CNN) has been achieving the highest accuracy, whereas the one-stage approach (\eg, SSD) has the advantage of high efficiency. To inherit the merits of both while overcoming their disadvantages, in this paper, we propose a novel single-shot based detector, called RefineDet, that achieves better accuracy than two-stage methods and maintains comparable efficiency of one-stage methods. RefineDet consists of two inter-connected modules, namely, the anchor refinement module and the object detection module. Specifically, the former aims to (1) filter out negative anchors to reduce search space for the classifier, and (2) coarsely adjust the locations and sizes of anchors to provide better initialization for the subsequent regressor. The latter module takes the refined anchors as the input from the former to further improve the regression and predict multi-class label. Meanwhile, we design a transfer connection block to transfer the features in the anchor refinement module to predict locations, sizes and class labels of objects in the object detection module. The multi-task loss function enables us to train the whole network in an end-to-end way. Extensive experiments on PASCAL VOC 2007, PASCAL VOC 2012, and MS COCO demonstrate that RefineDet achieves state-of-the-art detection accuracy with high efficiency. Code is available at \url{https://github.com/sfzhang15/RefineDet}.

\end{abstract}

%%%%%%%%% BODY TEXT
\section{Introduction}

Object detection has achieved significant advances in recent years, with the framework of deep neural networks (DNN). The current DNN detectors of state-of-the-art can be divided into two categories: (1) the two-stage approach, including \cite{DBLP:conf/eccv/CaiFFV16,DBLP:conf/iccv/Girshick15,DBLP:journals/pami/RenHG017,DBLP:conf/cvpr/ShrivastavaGG16}, and (2) the one-stage approach, including \cite{DBLP:conf/eccv/LiuAESRFB16, DBLP:journals/corr/RedmonF16}. In the two-stage approach, a sparse set of candidate object boxes is first generated, and then they are further classified and regressed. The two-stage methods have been achieving top performances on several challenging benchmarks, including PASCAL VOC \cite{DBLP:journals/ijcv/EveringhamGWWZ10} and MS COCO \cite{DBLP:conf/eccv/LinMBHPRDZ14}.

The one-stage approach detects objects by regular and dense sampling over locations, scales and aspect ratios. The main advantage of this is its high computational efficiency. However, its detection accuracy is usually behind that of the two-stage approach, one of the main reasons being due to the class imbalance problem \cite{DBLP:conf/iccv/LinPRK17}.

Some recent methods in the one-stage approach aim to address the class imbalance problem, to improve the detection accuracy. Kong \etal \cite{DBLP:conf/cvpr/KongSYLLC17} use the objectness prior constraint on convolutional feature maps to significantly reduce the search space of objects. Lin \etal \cite{DBLP:conf/iccv/LinPRK17} address the class imbalance issue by reshaping the standard cross entropy loss to focus training on a sparse set of hard examples and down-weights the loss assigned to well-classified examples. Zhang \etal \cite{DBLP:conf/iccv/abs-1708-05237} design a max-out labeling mechanism to reduce false positives resulting from class imbalance.

\begin{figure*}[t]
\centering
\includegraphics[width=0.95\linewidth]{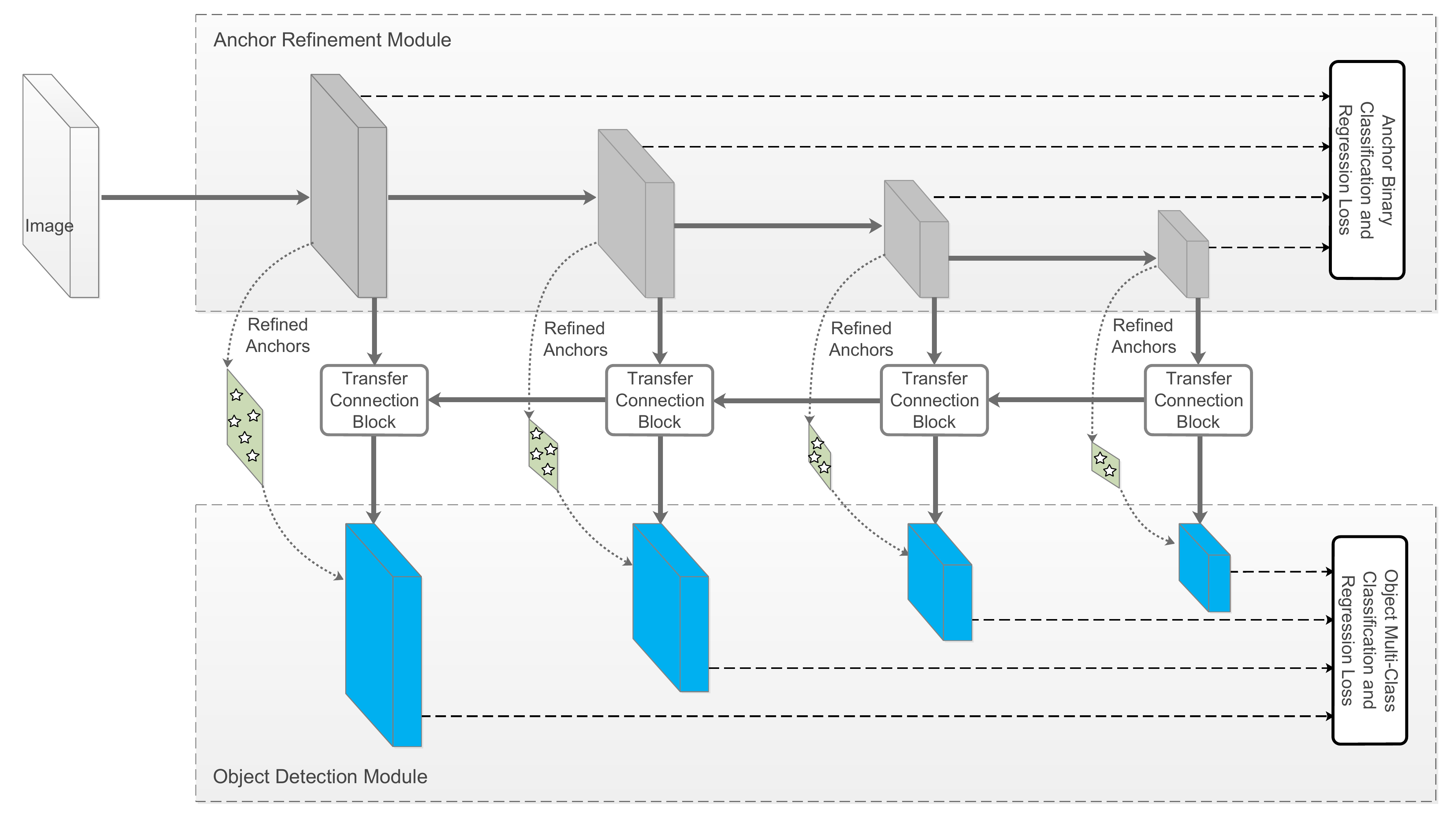}
\vspace{-2mm}
\caption{Architecture of RefineDet. For better visualization, we only display the layers used for detection. The celadon parallelograms denote the refined anchors associated with different feature layers. The stars represent the centers of the refined anchor boxes, which are not regularly paved on the image.}
\label{fig:net-structure}
\end{figure*}

In our opinion, the current state-of-the-art two-stage methods, \eg, Faster R-CNN \cite{DBLP:journals/pami/RenHG017}, R-FCN \cite{DBLP:conf/nips/DaiLHS16}, and FPN \cite{DBLP:conf/cvpr/LinDGHHB17}, have three advantages over the one-stage methods as follows: (1) using two-stage structure with sampling heuristics to handle class imbalance; (2) using two-step cascade to regress the object box parameters; (3) using two-stage features to describe the objects\footnote{In case of Faster R-CNN, the features (excluding shared features) in the first stage (\ie, RPN) are trained for the binary classification (being an object or not), while the features (excluding shared features) in the second stage(\ie, Fast R-CNN) are trained for the multi-class classification (background or object classes).}. In this work, we design a novel object detection framework, called RefineDet, to inherit the merits of the two approaches (\ie, one-stage and two-stage approaches) and overcome their shortcomings. It improves the architecture of the one-stage approach, by using two inter-connected modules (see Figure \ref{fig:net-structure}), namely, the anchor \footnote{We denote the reference bounding box as ``anchor box'', which is also called ``anchor'' for simplicity, as in \cite{DBLP:journals/pami/RenHG017}. However, in \cite{DBLP:conf/eccv/LiuAESRFB16}, it is called ``default box''.} refinement module (ARM) and the object detection module (ODM). Specifically, the ARM is designed to (1) identify and remove negative anchors to reduce search space for the classifier, and (2) coarsely adjust the locations and sizes of anchors to provide better initialization for the subsequent regressor. The ODM takes the refined anchors as the input from the former to further improve the regression and predict multi-class labels. As shown in Figure \ref{fig:net-structure}, these two inter-connected modules imitate the two-stage structure and thus inherit the three aforementioned advantages to produce accurate detection results with high efficiency. In addition, we design a transfer connection block (TCB) to transfer the features\footnote{The features in the ARM focus on distinguishing positive anchors from background. We design the TCB to transfer the features in the ARM to handle the more challenging tasks in the ODM, \ie, predict accurate object locations, sizes and multi-class labels.} in the ARM to predict locations, sizes, and class labels of objects in the ODM. The multi-task loss function enables us to train the whole network in an end-to-end way.

Extensive experiments on PASCAL VOC 2007, PASCAL VOC 2012, and MS COCO benchmarks demonstrate that RefineDet outperforms the state-of-the-art methods. Specifically, it achieves $85.8\%$ and $86.8\%$ mAPs on VOC 2007 and 2012, with VGG-16 network. Meanwhile, it outperforms the previously best published results from both one-stage and two-stage approaches by achieving $41.8\%$ AP\footnote{Based on the evaluation protocol in MS COCO \cite{DBLP:conf/eccv/LinMBHPRDZ14}, AP is the single most important metric, which is computed by averaging over all $10$ intersection over union (IoU) thresholds (\ie, in the range [$0.5$:$0.95$] with uniform step size $0.05$) of $80$ categories.} on MS COCO {\tt test-dev} with ResNet-101. In addition, RefineDet is time efficient, \ie, it runs at $40.2$ FPS and $24.1$ FPS on a NVIDIA Titan X GPU with the input sizes $320\times320$ and $512\times512$ in inference.

The main contributions of this work are summarized as follows. (1) We introduce a novel one-stage framework for object detection, composed of two inter-connected modules, \ie, the ARM and the ODM. This leads to performance better than the two-stage approach while maintaining high efficiency of the one-stage approach. (2) To ensure the effectiveness, we design the TCB to transfer the features in the ARM to handle more challenging tasks, \ie, predict accurate object locations, sizes and class labels, in the ODM. (3) RefineDet achieves the latest state-of-the-art results on generic object detection (\ie, PASCAL VOC 2007 \cite{pascal-voc-2007}, PASCAL VOC 2012 \cite{pascal-voc-2012} and MS COCO \cite{DBLP:conf/eccv/LinMBHPRDZ14}).

\section{Related Work}

{\noindent \textbf{Classical Object Detectors.}} Early object detection methods are based on the sliding-window paradigm, which apply the hand-crafted features and classifiers on dense image grids to find objects. As one of the most successful methods, Viola and Jones \cite{DBLP:conf/cvpr/ViolaJ01} use Haar feature and AdaBoost to train a series of cascaded classifiers for face detection, achieving satisfactory accuracy with high efficiency. DPM \cite{DBLP:journals/pami/FelzenszwalbGMR10} is another popular method using mixtures of multi-scale deformable part models to represent highly variable object classes, maintaining top results on PASCAL VOC \cite{DBLP:journals/ijcv/EveringhamGWWZ10} for many years. However, with the arrival of deep convolutional network, the object detection task is quickly dominated by the CNN-based detectors, which can be roughly divided into two categories, \ie, the two-stage approach and one-stage approach.

{\noindent \textbf{Two-Stage Approach.}} The two-stage approach consists of two parts, where the first one (\eg, Selective Search \cite{DBLP:journals/ijcv/UijlingsSGS13}, EdgeBoxes \cite{DBLP:conf/eccv/ZitnickD14}, DeepMask \cite{DBLP:conf/nips/PinheiroCD15,DBLP:conf/eccv/PinheiroLCD16}, RPN \cite{DBLP:journals/pami/RenHG017}) generates a sparse set of candidate object proposals, and the second one determines the accurate object regions and the corresponding class labels using convolutional networks. Notably, the two-stage approach (\eg, R-CNN \cite{DBLP:conf/cvpr/GirshickDDM14}, SPPnet \cite{DBLP:conf/eccv/HeZR014}, Fast R-CNN \cite{DBLP:conf/iccv/Girshick15} to Faster R-CNN \cite{DBLP:journals/pami/RenHG017}) achieves dominated performance on several challenging datasets (\eg, PASCAL VOC 2012 \cite{pascal-voc-2012} and MS COCO \cite{DBLP:conf/eccv/LinMBHPRDZ14}). After that, numerous effective techniques are proposed to further improve the performance, such as architecture diagram \cite{DBLP:conf/nips/DaiLHS16,DBLP:journals/corr/LeeEK17,DBLP:conf/iccv/abs-1708-02863}, training strategy \cite{DBLP:conf/cvpr/ShrivastavaGG16,DBLP:conf/cvpr/WangSG17}, contextual reasoning \cite{DBLP:conf/cvpr/BellZBG16,DBLP:conf/iccv/GidarisK15,DBLP:conf/eccv/ShrivastavaG16,DBLP:conf/eccv/ZengOYYW16} and multiple layers exploiting \cite{DBLP:conf/eccv/CaiFFV16,DBLP:conf/cvpr/KongYCS16,DBLP:conf/cvpr/LinDGHHB17,DBLP:journals/corr/ShrivastavaSMG16}.

{\noindent \textbf{One-Stage Approach.}} Considering the high efficiency, the one-stage approach attracts much more attention recently. Sermanet \etal \cite{DBLP:journals/corr/SermanetEZMFL13} present the OverFeat method for classification, localization and detection based on deep ConvNets, which is trained end-to-end, from raw pixels to ultimate categories. Redmon \etal \cite{DBLP:conf/cvpr/RedmonDGF16} use a single feed-forward convolutional network to directly predict object classes and locations, called YOLO, which is extremely fast. After that, YOLOv2 \cite{DBLP:journals/corr/RedmonF16} is proposed to improve YOLO in several aspects, \ie, add batch normalization on all convolution layers, use high resolution classifier, use convolution layers with anchor boxes to predict bounding boxes instead of the fully connected layers, etc. Liu \etal \cite{DBLP:conf/eccv/LiuAESRFB16} propose the SSD method, which spreads out anchors of different scales to multiple layers within a ConvNet and enforces each layer to focus on predicting objects of a certain scale. DSSD \cite{DBLP:journals/corr/FuLRTB17} introduces additional context into SSD via deconvolution to improve the accuracy. DSOD \cite{DBLP:conf/iccv/abs-1708-01241} designs an efficient framework and a set of principles to learn object detectors from scratch, following the network structure of SSD. To improve the accuracy, some one-stage methods \cite{DBLP:conf/cvpr/KongSYLLC17,DBLP:conf/iccv/LinPRK17,DBLP:conf/iccv/abs-1708-05237} aim to address the extreme class imbalance problem by re-designing the loss function or classification strategies. Although the one-stage detectors have made good progress, their accuracy still trails that of two-stage methods.

\section{Network Architecture}
Refer to the overall network architecture shown in Figure \ref{fig:net-structure}. Similar to SSD \cite{DBLP:conf/eccv/LiuAESRFB16}, RefineDet is based on a feed-forward convolutional network that produces a fixed number of bounding boxes and the scores indicating the presence of different classes of objects in those boxes, followed by the non-maximum suppression to produce the final result. RefineDet is formed by two inter-connected modules, \ie, the ARM and the ODM. The ARM aims to remove negative anchors so as to reduce search space for the classifier and also coarsely adjust the locations and sizes of anchors to provide better initialization for the subsequent regressor, whereas ODM aims to regress accurate object locations and predict multi-class labels based on the refined anchors. The ARM is constructed by removing the classification layers and adding some auxiliary structures of two base networks (\ie, VGG-16 \cite{DBLP:journals/corr/SimonyanZ14a} and ResNet-101 \cite{DBLP:conf/cvpr/HeZRS16} pretrained on ImageNet \cite{DBLP:journals/ijcv/RussakovskyDSKS15}) to meet our needs. The ODM is composed of the outputs of TCBs followed by the prediction layers (\ie, the convolution layers with $3\times3$ kernel size), which generates the scores for object classes and shape offsets relative to the refined anchor box coordinates. The following explain three core components in RefineDet, \ie, (1) transfer connection block (TCB), converting the features from the ARM to the ODM for detection; (2) two-step cascaded regression, accurately regressing the locations and sizes of objects; (3) negative anchor filtering, early rejecting well-classified negative anchors and mitigate the imbalance issue.

%connects the prediction layers (\ie, the convolution layers with $3\times3$ kernel size) to the output of TCBs for detection, \ie, generates the scores for object classes and shape offsets relative to the anchor box coordinates

\begin{figure}
\centering
\includegraphics[width=0.5\linewidth]{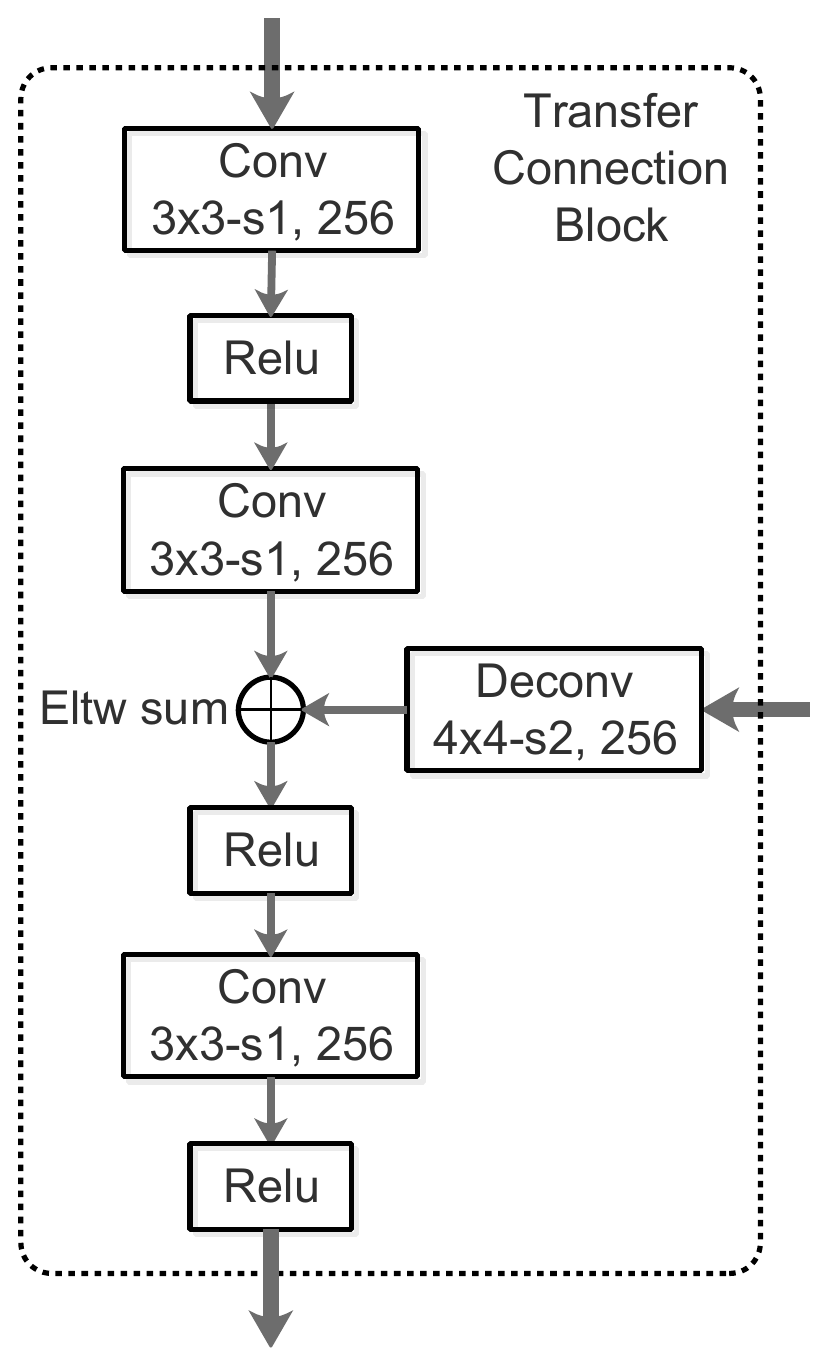}
\vspace{-2mm}
\caption{The overview of the transfer connection block.}
\label{fig:transfer-connection-block}
%\vspace{-2mm}
\end{figure}

{\noindent \textbf{Transfer Connection Block.}}
To link between the ARM and ODM, we introduce  the TCBs to convert features of different layers from the ARM, into the form required by the ODM, so that the ODM can share features from the ARM. Notably, from the ARM, we only use the TCBs on the feature maps associated with anchors. Another function of the TCBs is to integrate large-scale context \cite{DBLP:journals/corr/FuLRTB17,DBLP:conf/cvpr/LinDGHHB17} by adding the high-level features to the transferred features to improve detection accuracy. To match the dimensions between them, we use the deconvolution operation to enlarge the high-level feature maps and sum them in the element-wise way. Then, we add a convolution layer after the summation to ensure the discriminability of features for detection. The architecture of the TCB is shown in Figure \ref{fig:transfer-connection-block}.

{\noindent \textbf{Two-Step Cascaded Regression.}}
Current one-stage methods \cite{DBLP:journals/corr/FuLRTB17,DBLP:conf/cvpr/KongSYLLC17,DBLP:conf/eccv/LiuAESRFB16} rely on one-step regression based on various feature layers with different scales to predict the locations and sizes of objects, which is rather inaccurate in some challenging scenarios, especially for the small objects. To that end, we present a two-step cascaded regression strategy to regress the locations and sizes of objects. That is, we use the ARM to first adjust the locations and sizes of anchors to provide better initialization for the regression in the ODM. Specifically, we associate $n$ anchor boxes with each regularly divided cell on the feature map. The initial position of each anchor box relative to its corresponding cell is fixed. At each feature map cell, we predict four offsets of the refined anchor boxes relative to the original tiled anchors and two confidence scores indicating the presence of foreground objects in those boxes. Thus, we can yield $n$ refined anchor boxes at each feature map cell.

After obtaining the refined anchor boxes, we pass them to the corresponding feature maps in the ODM to further generate object categories and accurate object locations and sizes, as shown in Figure \ref{fig:net-structure}. The corresponding feature maps in the ARM and the ODM have the same dimension. We calculate $c$ class scores and the four accurate offsets of objects relative to the refined anchor boxes, yielding $c+4$ outputs for each refined anchor boxes to complete the detection task. This process is similar to the default boxes used in SSD \cite{DBLP:conf/eccv/LiuAESRFB16}. However, in contrast to SSD \cite{DBLP:conf/eccv/LiuAESRFB16} directly uses the regularly tiled default boxes for detection, RefineDet uses two-step strategy, \ie, the ARM generates the refined anchor boxes, and the ODM takes the refined anchor boxes as input for further detection, leading to more accurate detection results, especially for the small objects.

{\noindent \textbf{Negative Anchor Filtering.}}
To early reject well-classified negative anchors and mitigate the imbalance issue, we design a negative anchor filtering mechanism. Specifically, in training phase, for a refined anchor box, if its negative confidence is larger than a preset threshold $\theta$ (\ie, set $\theta=0.99$ empirically), we will discard it in training the ODM. That is, we only pass the refined hard negative anchor boxes and refined positive anchor boxes to train the ODM. Meanwhile, in the inference phase, if a refined anchor box is assigned with a negative confidence larger than $\theta$, it will be discarded in the ODM for detection.

\section{Training and Inference}

{\noindent \textbf{Data Augmentation.}}
We use several data augmentation strategies presented in \cite{DBLP:conf/eccv/LiuAESRFB16} to construct a robust model to adapt to variations of objects. That is, we randomly expand and crop the original training images with additional random photometric distortion \cite{DBLP:journals/corr/Howard13} and flipping to generate the training samples. Please refer to \cite{DBLP:conf/eccv/LiuAESRFB16} for more details.

{\noindent \textbf{Backbone Network.}}
We use VGG-16 \cite{DBLP:journals/corr/SimonyanZ14a} and ResNet-101 \cite{DBLP:conf/cvpr/HeZRS16} as the backbone networks in our RefineDet, which are pretrained on the ILSVRC CLS-LOC dataset \cite{DBLP:journals/ijcv/RussakovskyDSKS15}. Notably, RefineDet can also work on other pretrained networks, such as Inception V2 \cite{DBLP:conf/icml/IoffeS15}, Inception ResNet \cite{DBLP:conf/aaai/SzegedyIVA17}, and ResNeXt-101 \cite{DBLP:journals/corr/XieGDTH16}. Similar to DeepLab-LargeFOV \cite{DBLP:conf/iclr/ChenPKMY14}, we convert fc6 and fc7 of VGG-16 to convolution layers conv\_fc6 and conv\_fc7 via subsampling parameters. Since conv4\_3 and conv5\_3 have different feature scales compared to other layers, we use L2 normalization \cite{DBLP:conf/iclrw/LiuRB15} to scale the feature norms in conv4\_3 and conv5\_3 to $10$ and $8$, then learn the scales during back propagation. Meanwhile, to capture high-level information and drive object detection at multiple scales, we also add two extra convolution layers (\ie, conv6\_1 and conv6\_2) to the end of the truncated VGG-16 and one extra residual block (\ie, res6) to the end of the truncated ResNet-101, respectively.

{\noindent \textbf{Anchors Design and Matching.}}
To handle different scales of objects, we select four feature layers with the total stride sizes $8$, $16$, $32$, and $64$ pixels for both VGG-16 and ResNet-101\footnote{For the VGG-16 base network, the conv4\_3, conv5\_3, conv\_fc7, and conv6\_2 feature layers are used to predict the locations, sizes and confidences of objects. While for the ResNet-101 base network, res3b3, res4b22, res5c, and res6 are used for prediction.}, associated with several different scales of anchors for prediction. Each feature layer is associated with one specific scale of anchors (\ie, the scale is $4$ times of the total stride size of the corresponding layer) and three aspect ratios (\ie, $0.5$, $1.0$, and $2.0$). We follow the design of anchor scales over different layers in \cite{DBLP:conf/iccv/abs-1708-05237}, which ensures that different scales of anchors have the same tiling density \cite{DBLP:conf/ccbr/ZhangZLSWL17,DBLP:conf/ijcb/abs-1708-05234} on the image. Meanwhile, during the training phase, we determine the correspondence between the anchors and ground truth boxes based on the jaccard overlap \cite{DBLP:conf/cvpr/ErhanSTA14}, and train the whole network end-to-end accordingly. Specifically, we first match each ground truth to the anchor box with the best overlap score, and then match the anchor boxes to any ground truth with overlap higher than $0.5$.

{\noindent \textbf{Hard Negative Mining.}}
After matching step, most of the anchor boxes are negatives, even for the ODM, where some easy negative anchors are rejected by the ARM. Similar to SSD \cite{DBLP:conf/eccv/LiuAESRFB16}, we use hard negative mining to mitigate the extreme foreground-background class imbalance, \ie, we select some negative anchor boxes with top loss values to make the ratio between the negatives and positives below $3:1$, instead of using all negative anchors or randomly selecting the negative anchors in training.

{\noindent \textbf{Loss Function.}}
The loss function for RefineDet consists of two parts, \ie, the loss in the ARM and the loss in the ODM. For the ARM, we assign a binary class label (of being an object or not) to each anchor and regress its location and size simultaneously to get the refined anchor. After that, we pass the refined anchors with the negative confidence less than the threshold to the ODM to further predict object categories and accurate object locations and sizes. With these definitions, we define the loss function as:
\begin{equation}
\begin{array}{cl}
&{\cal L}(\{p_i\},\{x_i\},\{c_i\},\{t_i\})=\frac{1}{N_{\text{arm}}} \big( \sum_{i}{\cal L}_{\text{b}}(p_i,[l_i^\ast\geq1])  \\
&+\sum_{i}[l_i^\ast\geq1]{\cal L}_{\text{r}}(x_i, g_i^\ast)\big) + \frac{1}{N_{\text{odm}}} \big( \sum_{i}{\cal L}_{\text{m}}(c_i, l_i^\ast)  \\
&+\sum_{i}[l_i^\ast\geq1]{\cal L}_{\text{r}}(t_i, g_i^\ast)\big)
\end{array}
\label{1}
\end{equation}
where $i$ is the index of anchor in a mini-batch, $l_i^\ast$ is the ground truth class label of anchor $i$, $g_i^\ast$ is the ground truth location and size of anchor $i$. $p_i$ and $x_i$ are the predicted confidence of the anchor $i$ being an object and refined coordinates of the anchor $i$ in the ARM. $c_i$ and $t_i$ are the predicted object class and coordinates of the bounding box in the ODM. $N_{\text{arm}}$ and $N_{\text{odm}}$ are the numbers of positive anchors in the ARM and ODM, respectively. The binary classification loss ${\cal L}_{\text{b}}$ is the cross-entropy/log loss over two classes (object {\em vs.} not object), and the multi-class classification loss ${\cal L}_{\text{m}}$ is the softmax loss over multiple classes confidences. Similar to Fast R-CNN \cite{DBLP:conf/iccv/Girshick15}, we use the smooth L1 loss as the regression loss $L_{\text{r}}$. The Iverson bracket indicator function $[l_i^\ast\geq1]$ outputs $1$ when the condition is true, \ie, $l_i^\ast\geq1$ (the anchor is not the negative), and $0$ otherwise. Hence $[l_i^\ast\geq1]{\cal L}_{\text{r}}$ indicates that the regression loss is ignored for negative anchors. Notably, if $N_{\text{arm}}=0$, we set ${\cal L}_{\text{b}}(p_i,[l_i^\ast\geq1])=0$ and ${\cal L}_{\text{r}}(x_i, g_i^\ast)=0$; and if $N_{\text{odm}}=0$, we set ${\cal L}_{\text{m}}(c_i, l_i^\ast)=0$ and ${\cal L}_{\text{r}}(t_i, g_i^\ast)=0$ accordingly.

{\noindent \textbf{Optimization.}}
As mentioned above, the backbone network (\eg, VGG-16 and ResNet-101) in our RefineDet method is pretrained on the ILSVRC CLS-LOC dataset \cite{DBLP:journals/ijcv/RussakovskyDSKS15}. We use the ``xavier'' method \cite{DBLP:journals/jmlr/GlorotB10} to randomly initialize the parameters in the two extra added convolution layers (\ie, conv6\_1 and conv6\_2) of VGG-16 based RefineDet, and draw the parameters from a zero-mean Gaussian distribution with standard deviation $0.01$ for the extra residual block (\ie, res6) of ResNet-101 based RefineDet. We set the default batch size to $32$ in training. Then, the whole network is fine-tuned using SGD with $0.9$ momentum and $0.0005$ weight decay. We set the initial learning rate to $10^{-3}$, and use slightly different learning rate decay policy for different dataset, which will be described in details later.

\begin{table*}[t]
\centering
\caption{Detection results on PASCAL VOC dataset. For VOC 2007, all methods are trained on VOC 2007 and VOC 2012 {\tt trainval} sets and tested on VOC 2007 {\tt test} set. For VOC 2012, all methods are trained on VOC 2007 and VOC 2012 {\tt trainval} sets plus VOC 2007 {\tt test} set, and tested on VOC 2012 {\tt test} set. Bold fonts indicate the best mAP.}
\footnotesize \setlength{\tabcolsep}{2.5pt}
\begin{tabular}{p{3.0cm}<{\centering}|p{3.0cm}<{\centering}|p{2.5cm}<{\centering}|p{1.5cm}<{\centering}|p{1.5cm}<{\centering}|p{2.2cm}<{\centering}|p{2.2cm}<{\centering}}
\toprule[1.5pt]
\multirow{2}{*}{Method} &\multirow{2}{*}{Backbone} &\multirow{2}{*}{Input size} &\multirow{2}{*}{$\#$Boxes} &\multirow{2}{*}{FPS} &\multicolumn{2}{c}{mAP (\%)} \\
\cline{6-7}
& & & & &VOC 2007 &VOC 2012 \\
\hline
\textit{two-stage:} & & & & & &\\
Fast R-CNN\cite{DBLP:conf/iccv/Girshick15}     &VGG-16 &$\sim1000\times600$ &$\sim2000$ &0.5 &70.0 &68.4\\
Faster R-CNN\cite{DBLP:journals/pami/RenHG017} &VGG-16 &$\sim1000\times600$ &300 &7 &73.2 &70.4\\
OHEM\cite{DBLP:conf/cvpr/ShrivastavaGG16}      &VGG-16 &$\sim1000\times600$ &300 &7 &74.6 &71.9\\
HyperNet\cite{DBLP:conf/cvpr/KongYCS16}        &VGG-16 &$\sim1000\times600$ &100 &0.88 &76.3 &71.4\\
Faster R-CNN\cite{DBLP:journals/pami/RenHG017} &ResNet-101 &$\sim1000\times600$ &300 &2.4 &76.4 & 73.8\\
ION\cite{DBLP:conf/cvpr/BellZBG16}             &VGG-16 &$\sim1000\times600$ &4000 &1.25 &76.5 &76.4\\
MR-CNN\cite{DBLP:conf/iccv/GidarisK15}         &VGG-16 &$\sim1000\times600$ &250 &0.03 &78.2 &73.9\\
R-FCN\cite{DBLP:conf/nips/DaiLHS16}            &ResNet-101 &$\sim1000\times600$ &300 &9 &80.5 &77.6\\
CoupleNet\cite{DBLP:conf/iccv/abs-1708-02863}  &ResNet-101 &$\sim1000\times600$ &300 &8.2 &82.7 &80.4\\
\hline
\hline
\textit{one-stage:} & & & & & &\\
YOLO\cite{DBLP:conf/cvpr/RedmonDGF16}       &GoogleNet~\cite{DBLP:conf/cvpr/SzegedyLJSRAEVR15} &$448\times448$ &98 &45 &63.4 &57.9\\
RON384\cite{DBLP:conf/cvpr/KongSYLLC17}     &VGG-16 &$384\times384$ &30600 &15 &75.4 &73.0\\
SSD321\cite{DBLP:journals/corr/FuLRTB17}    &ResNet-101 &$321\times321$ &17080 &11.2 &77.1 &75.4\\
SSD300$^*$\cite{DBLP:conf/eccv/LiuAESRFB16} &VGG-16 &$300\times300$ &8732 &46 &77.2 &75.8\\
DSOD300\cite{DBLP:conf/iccv/abs-1708-01241} &DS/64-192-48-1 &$300\times300$ &8732 &17.4 &77.7 &76.3\\
YOLOv2\cite{DBLP:journals/corr/RedmonF16}   &Darknet-19 &$544\times544$ &845 &40 &78.6 &73.4\\
DSSD321\cite{DBLP:journals/corr/FuLRTB17}   &ResNet-101 &$321\times321$ &17080 &9.5 &78.6 &76.3\\
SSD512$^*$\cite{DBLP:conf/eccv/LiuAESRFB16} &VGG-16 &$512\times512$ &24564 &19 &79.8 &78.5\\
SSD513\cite{DBLP:journals/corr/FuLRTB17}    &ResNet-101 &$513\times513$ &43688 &6.8 &80.6 &79.4\\
DSSD513\cite{DBLP:journals/corr/FuLRTB17}   &ResNet-101 &$513\times513$ &43688 &5.5 &81.5 &80.0\\
\hline
RefineDet320    &VGG-16     &$320\times320$ &6375  &40.3 &80.0 &78.1\\
RefineDet512    &VGG-16     &$512\times512$ &16320 &24.1 &81.8 &80.1\\
RefineDet320+   &VGG-16     &-  &- &- &83.1 &82.7\\
RefineDet512+   &VGG-16     &-  &- &- &{\bf 83.8} &{\bf 83.5}\\
\bottomrule[1.5pt]
\end{tabular}
\label{tab:pascal-voc}
%\vspace{-2mm}
\end{table*}

{\noindent \textbf{Inference.}}
At inference phase, the ARM first filters out the regularly tiled anchors with the negative confidence scores larger than the threshold $\theta$, and then refines the locations and sizes of remaining anchors. After that, the ODM takes over these refined anchors, and outputs top $400$ high confident detections per image. Finally, we apply the non-maximum suppression with jaccard overlap of $0.45$ per class and retain the top $200$ high confident detections per image to produce the final detection results.

\section{Experiments}
Experiments are conducted on three datasets: PASCAL VOC 2007, PASCAL VOC 2012 and MS COCO. The PASCAL VOC and MS COCO datasets include $20$ and $80$ object classes, respectively. The classes in PASCAL VOC are the subset of that in MS COCO. We implement RefineDet in Caffe \cite{DBLP:conf/mm/JiaSDKLGGD14}. All the training and testing codes and the trained models are available at \url{https://github.com/sfzhang15/RefineDet}.

\subsection{PASCAL VOC 2007}
All models are trained on the VOC 2007 and VOC 2012 {\tt trainval} sets, and tested on the VOC 2007 {\tt test} set. We set the learning rate to $10^{-3}$ for the first $80k$ iterations, and decay it to $10^{-4}$ and $10^{-5}$ for training another $20k$ and $20k$ iterations, respectively. We use the default batch size $32$ in training, and only use VGG-16 as the backbone network for all the experiments on the PASCAL VOC dataset, including VOC 2007 and VOC 2012.

We compare RefineDet\footnote{Due to the shortage of computational resources, we only train RefineDet with two kinds of input size, \ie, $320\times320$ and $512\times512$. We believe the accuracy of RefineDet can be further improved using larger input images.} with the state-of-the-art detectors in Table \ref{tab:pascal-voc}. With low dimension input (\ie, $320\times320$), RefineDet produces $80.0\%$ mAP without bells and whistles, which is the first method achieving above $80\%$ mAP with such small input images, much better than several modern objectors. By using larger input size $512\times512$, RefineDet achieves $81.8\%$ mAP, surpassing all one-stage methods, \eg, RON384 \cite{DBLP:conf/cvpr/KongSYLLC17}, SSD513 \cite{DBLP:journals/corr/FuLRTB17}, DSSD513 \cite{DBLP:journals/corr/FuLRTB17}, etc. Comparing to the two-stage methods, RefineDet512 performs better than most of them except CoupleNet \cite{DBLP:conf/iccv/abs-1708-02863}, which is based on ResNet-101 and uses larger input size (\ie, $\sim1000\times600$) than our RefineDet512. As pointed out in \cite{DBLP:conf/cvpr/HuangRSZKFFWSG016}, the input size significantly influences detection accuracy. The reason is that high resolution inputs make the detectors ``seeing'' small objects clearly to increase successful detections. To reduce the impact of input size for a fair comparison, we use the multi-scale testing strategy to evaluate RefineDet, achieving $83.1\%$ (RefineDet320+) and $83.8\%$ (RefineDet512+) mAPs, which are much better than the state-of-the-art methods.

\subsubsection{Run Time Performance}
We present the inference speed of RefineDet and the state-of-the-art methods in the fifth column of Table \ref{tab:pascal-voc}. The speed is evaluated with batch size $1$ on a machine with NVIDIA Titan X, CUDA 8.0 and cuDNN v6. As shown in Table \ref{tab:pascal-voc}, we find that RefineDet processes an image in $24.8$ms ($40.3$ FPS) and $41.5$ms ($24.1$ FPS) with input sizes $320\times320$ and $512\times512$, respectively. To the best of our knowledge, RefineDet is the first real-time method to achieve detection accuracy above $80\%$ mAP on PASCAL VOC 2007. Comparing to SSD, RON, DSSD and DSOD, RefineDet associates fewer anchor boxes on the feature maps (\eg, $24564$ anchor boxes in SSD512$^*$\cite{DBLP:conf/eccv/LiuAESRFB16} {\it vs.} $16320$ anchor boxes in RefineDet512). However, RefineDet still achieves top accuracy with high efficiency, mainly thanks to the design of two inter-connected modules, (\eg, two-step regression), which enables RefineDet to adapt to different scales and aspect ratios of objects. Meanwhile, only YOLO and SSD300$^*$ are slightly faster than our RefineDet320, but their accuracy are $16.6\%$ and $2.5\%$ worse than ours. In summary, RefineDet achieves the best trade-off between accuracy and speed.

\begin{table*}[t]
\centering
\caption{Detection results on MS COCO {\tt test-dev} set. Bold fonts indicate the best performance.}
%\vspace{-2mm}
\footnotesize \setlength{\tabcolsep}{9.5pt}
\begin{threeparttable}
\begin{tabular}{c|c|c|ccc|ccc}
\toprule[1.5pt]
Method &Data &Backbone &AP &AP$_{50}$ &AP$_{75}$ &AP$_{\it S}$ &AP$_{\it M}$ &AP$_{\it L}$\\
\hline
\textit{two-stage:} & & & & & & & & \\
Fast R-CNN \cite{DBLP:conf/iccv/Girshick15} &train &VGG-16 &19.7 &35.9 &- &- &- &- \\
Faster R-CNN \cite{DBLP:journals/pami/RenHG017} &trainval &VGG-16 &21.9 &42.7 &- &- &- &- \\
OHEM \cite{DBLP:conf/cvpr/ShrivastavaGG16} &trainval &VGG-16 &22.6 &42.5 &22.2 &5.0 &23.7 &37.9 \\
ION \cite{DBLP:conf/cvpr/BellZBG16} &train &VGG-16 &23.6 &43.2 &23.6 &6.4 &24.1 &38.3\\
OHEM++ \cite{DBLP:conf/cvpr/ShrivastavaGG16} &trainval &VGG-16 &25.5 &45.9 &26.1 &7.4 &27.7 &40.3 \\
R-FCN \cite{DBLP:conf/nips/DaiLHS16} &trainval &ResNet-101 &29.9 &51.9 &- &10.8 &32.8 &45.0\\
CoupleNet \cite{DBLP:conf/iccv/abs-1708-02863} &trainval &ResNet-101 &34.4 &54.8 &37.2 &13.4 &38.1 &50.8 \\
Faster R-CNN by G-RMI \cite{DBLP:conf/cvpr/HuangRSZKFFWSG016} &- &Inception-ResNet-v2\cite{DBLP:conf/aaai/SzegedyIVA17} &34.7 &55.5 &36.7 &13.5 &38.1 &52.0 \\
Faster R-CNN+++ \cite{DBLP:conf/cvpr/HeZRS16} &trainval &ResNet-101-C4 &34.9 &55.7 &37.4 &15.6 &38.7 &50.9\\
Faster R-CNN w FPN \cite{DBLP:conf/cvpr/LinDGHHB17} &trainval35k &ResNet-101-FPN &36.2 &59.1 &39.0 &18.2 &39.0 &48.2 \\
Faster R-CNN w TDM \cite{DBLP:journals/corr/ShrivastavaSMG16} &trainval &Inception-ResNet-v2-TDM &36.8 &57.7 &39.2 &16.2 &39.8 &52.1 \\
Deformable R-FCN \cite{DBLP:conf/iccv/DaiQXLZHW17} &trainval &Aligned-Inception-ResNet &37.5 &58.0 &40.8 &19.4 &40.1 &52.5 \\
umd\_det \cite{DBLP:conf/iccv/BodlaSCD17}                         &trainval      &ResNet-101     &40.8 &62.4 &44.9 &23.0 &43.4 &53.2 \\
G-RMI \cite{DBLP:conf/cvpr/HuangRSZKFFWSG016} &trainval32k &Ensemble of Five Models &41.6 &61.9 &45.4 &23.9 &43.5 &54.9 \\
\hline
\hline
\textit{one-stage:} & & & & & & & & \\
YOLOv2 \cite{DBLP:journals/corr/RedmonF16} &trainval35k &DarkNet-19\cite{DBLP:journals/corr/RedmonF16} &21.6 &44.0 &19.2 &5.0 &22.4 &35.5\\
SSD300$^*$ \cite{DBLP:conf/eccv/LiuAESRFB16} &trainval35k &VGG-16 &25.1 &43.1 &25.8 &6.6 &25.9 &41.4\\
RON384++ \cite{DBLP:conf/cvpr/KongSYLLC17} &trainval &VGG-16 &27.4 &49.5 &27.1 &- &- &- \\
SSD321 \cite{DBLP:journals/corr/FuLRTB17} &trainval35k &ResNet-101 &28.0 &45.4 &29.3 &6.2 &28.3 &49.3\\
DSSD321 \cite{DBLP:journals/corr/FuLRTB17} &trainval35k &ResNet-101 &28.0 &46.1 &29.2 &7.4 &28.1 &47.6\\
SSD512$^*$ \cite{DBLP:conf/eccv/LiuAESRFB16} &trainval35k &VGG-16 &28.8 &48.5 &30.3 &10.9 &31.8 &43.5\\
SSD513 \cite{DBLP:journals/corr/FuLRTB17} &trainval35k &ResNet-101 &31.2 &50.4 &33.3 &10.2 &34.5 &49.8 \\
DSSD513 \cite{DBLP:journals/corr/FuLRTB17} &trainval35k &ResNet-101 &33.2 &53.3 &35.2 &13.0 &35.4 &51.1 \\
RetinaNet500 \cite{DBLP:conf/iccv/LinPRK17} &trainval35k &ResNet-101 &34.4 &53.1 &36.8 &14.7 &38.5 &49.1 \\
RetinaNet800 \cite{DBLP:conf/iccv/LinPRK17}\tnote{$\bm{\ast}$} &trainval35k &ResNet-101-FPN &39.1 &59.1 &42.3 &21.8 &42.7 &50.2 \\
\hline
RefineDet320  &trainval35k &VGG-16 &29.4 &49.2 &31.3 &10.0 &32.0 &44.4\\
RefineDet512  &trainval35k &VGG-16 &33.0 &54.5 &35.5 &16.3 &36.3 &44.3 \\
RefineDet320  &trainval35k &ResNet-101 &32.0 &51.4 &34.2 &10.5 &34.7 &50.4 \\
RefineDet512  &trainval35k &ResNet-101 &36.4 &57.5 &39.5 &16.6 &39.9 &51.4 \\
RefineDet320+ &trainval35k &VGG-16 &35.2 &56.1 &37.7 &19.5 &37.2 &47.0 \\
RefineDet512+ &trainval35k &VGG-16 &37.6 &58.7 &40.8 &22.7 &40.3 &48.3\\
RefineDet320+ &trainval35k &ResNet-101 &38.6 &59.9 &41.7 &21.1 &41.7 &52.3 \\
RefineDet512+ &trainval35k &ResNet-101 &{\bf 41.8} &{\bf 62.9} &{\bf 45.7} &{\bf 25.6} &{\bf 45.1} &{\bf 54.1} \\
\bottomrule[1.5pt]
\end{tabular}
\begin{tablenotes}
\item[$\bm{\ast}$] This entry reports the single model accuracy of RetinaNet method, trained with scale jitter and for $1.5\times$ longer than RetinaNet500.
\end{tablenotes}
\end{threeparttable}
\label{tab:coco}
\end{table*}

\begin{table}[t]
\centering
\caption{Effectiveness of various designs. All models are trained on VOC 2007 and VOC 2012 {\tt trainval} set and tested on VOC 2007 {\tt test} set.}
%\vspace{-2mm}
\footnotesize \setlength{\tabcolsep}{6.0pt}
\begin{tabular}{p{3.5cm}<{\centering}|p{0.7cm}<{\centering}p{0.7cm}<{\centering}p{0.7cm}<{\centering}p{0.7cm}<{\centering}}
\toprule[1.5pt]
\multicolumn{1}{c|}{Component}&\multicolumn{4}{c}{RefineDet320}\\
\hline
negative anchor filtering? & \Checkmark & & & \\
two-step cascaded regression? & \Checkmark & \Checkmark & & \\
transfer connection block? & \Checkmark & \Checkmark & \Checkmark &\\
\hline
mAP (\%) & 80.0 & 79.5 & 77.3 & 76.2\\
\bottomrule[1.5pt]
\end{tabular}
\label{tab:ablation}
%\vspace{-2mm}
\end{table}

\subsubsection{Ablation Study}
To demonstrate the effectiveness of different components in RefineDet, we construct four variants and evaluate them on VOC 2007, shown in Table \ref{tab:ablation}. Specifically, for a fair comparison, we use the same parameter settings and input size ($320\times320$) in evaluation. All models are trained on VOC 2007 and VOC 2012 {\tt trainval} sets, and tested on VOC 2007 {\tt test} set.

{\flushleft \textbf{Negative Anchor Filtering.}} To demonstrate the effectiveness of the negative anchor filtering, we set the confidence threshold $\theta$ of the anchors to be negative to $1.0$ in both training and testing. In this case, all refined anchors will be sent to the ODM for detection. Other parts of RefineDet remain unchanged. Removing negative anchor filtering leads to $0.5\%$ drop in mAP (\ie, $80.0\%$ {\it vs.} $79.5\%$). The reason is that most of these well-classified negative anchors will be filtered out during training, which solves the class imbalance issue to some extent.

{\flushleft \textbf{Two-Step Cascaded Regression.}} To validate the effectiveness of the two-step cascaded regression, we redesign the network structure by directly using the regularly paved anchors instead of the refined ones from the ARM (see the fourth column in Table \ref{tab:ablation}). As shown in Table \ref{tab:ablation}, we find that mAP is reduced from $79.5\%$ to $77.3\%$. This sharp decline (\ie, $2.2\%$) demonstrates that the two-step anchor cascaded regression significantly help promote the performance.

{\flushleft \textbf{Transfer Connection Block.}} We construct a network by cutting the TCBs in RefineDet and redefining the loss function in the ARM to directly detect multi-class of objects, just like SSD, to demonstrate the effect of the TCB. The detection accuracy of the model is presented in the fifth column in Table \ref{tab:ablation}. We compare the results in the fourth and fifth columns in Table \ref{tab:ablation} ($77.3\%$ {\it vs.} $76.2\%$) and find that the TCB improves the mAP by $1.1\%$. The main reason is that the model can inherit the discriminative features from the ARM, and integrate large-scale context information to improve the detection accuracy by using the TCB.

%We have proved the effectiveness of the components in our RefineDet method via subtle ablation experiments, but our main contribution is to propose this novel detection framework, which can be integrated into any algorithm to improve performance.

\subsection{PASCAL VOC 2012}
Following the protocol of VOC 2012, we submit the detection results of RefineDet to the public testing server for evaluation. We use VOC 2007 {\tt trainval} set and {\tt test} set plus VOC 2012 {\tt trainval} set ($21,503$ images) for training, and test on VOC 2012 {\tt test} set ($10,991$ images). We use the default batch size $32$ in training. Meanwhile, we set the learning rate to $10^{-3}$ in the first $160k$ iterations, and decay it to $10^{-4}$ and $10^{-5}$ for another $40k$ and $40k$ iterations.

Table \ref{tab:pascal-voc} shows the accuracy of the proposed RefineDet algorithm, as well as the state-of-the-art methods. Among the methods fed with input size $320\times320$, RefineDet320 obtains the top $78.1\%$ mAP, which is even better than most of those two-stage methods using about $1000\times600$ input size (\eg, $70.4\%$ mAP of Faster R-CNN \cite{DBLP:journals/pami/RenHG017} and $77.6\%$ mAP of R-FCN \cite{DBLP:conf/nips/DaiLHS16}). Using the input size $512\times512$, RefineDet improves mAP to $80.1\%$, which is surpassing all one-stage methods and only slightly lower than CoupleNet \cite{DBLP:conf/iccv/abs-1708-02863} (\ie, $80.4\%$). CoupleNet uses ResNet-101 as base network with $1000\times600$ input size. To reduce the impact of input size for a fair comparison, we also use multi-scale testing to evaluate RefineDet and obtain the state-of-the-art mAPs of $82.7\%$ (RefineDet320+) and $83.5\%$ (RefineDet512+).

\begin{table}
\centering
\caption{Detection results on PASCAL VOC dataset. All models are pre-trained on MS COCO, and fine-tuned on PASCAL VOC. Bold fonts indicate the best mAP.}
\vspace{-2mm}
\footnotesize \setlength{\tabcolsep}{1.5pt}
\begin{tabular}{c|c|c|c}
\toprule[1.5pt]
\multirow{2}{*}{Method}   &\multirow{2}{*}{Backbone} &\multicolumn{2}{c}{mAP (\%)} \\
\cline{3-4}
 & & VOC 2007 {\tt test} & VOC 2012 {\tt test}\\
\hline
\textit{two-stage:} & & &\\
Faster R-CNN\cite{DBLP:journals/pami/RenHG017}  &VGG-16 &78.8 &75.9   \\
OHEM++\cite{DBLP:conf/cvpr/ShrivastavaGG16}     &VGG-16 &- &80.1 \\
R-FCN\cite{DBLP:conf/nips/DaiLHS16}             &ResNet-101 &83.6 &82.0\\
\hline
\hline
\textit{one-stage:} & & &\\
SSD300\cite{DBLP:conf/eccv/LiuAESRFB16}         &VGG-16         &81.2 &79.3 \\
SSD512\cite{DBLP:conf/eccv/LiuAESRFB16}         &VGG-16         &83.2 &82.2 \\
RON384++\cite{DBLP:conf/cvpr/KongSYLLC17}       &VGG-16         &81.3 &80.7 \\
DSOD300\cite{DBLP:conf/iccv/abs-1708-01241}     &DS/64-192-48-1 &81.7 &79.3 \\
\hline
RefineDet320   &VGG-16        &84.0      &82.7 \\
RefineDet512   &VGG-16        &85.2      &85.0 \\
RefineDet320+  &VGG-16        &85.6      &86.0 \\
RefineDet512+  &VGG-16        &{\bf 85.8}      &{\bf 86.8}  \\
\bottomrule[1.5pt]
\end{tabular}
\label{tab:coco-to-voc}
\vspace{-2mm}
\end{table}

\subsection{MS COCO}
In addition to PASCAL VOC, we also evaluate RefineDet on MS COCO \cite{DBLP:conf/eccv/LinMBHPRDZ14}. Unlike PASCAL VOC, the detection methods using ResNet-101 always achieve better performance than those using VGG-16 on MS COCO. Thus, we also report the results of ResNet-101 based RefineDet. Following the protocol in MS COCO, we use the {\tt trainval35k} set \cite{DBLP:conf/cvpr/BellZBG16} for training and evaluate the results from {\tt test-dev} evaluation server. We set the batch size to $32$ in training\footnote{Due to the memory issue, we reduce the batch size to $20$ (which is the largest batch size we can use for training on a machine with $4$ NVIDIA M40 GPUs) to train the ResNet-101 based RefineDet with the input size $512\times512$, and train the model with $10^{-3}$ learning rate for the first $400k$ iterations, then $10^{-4}$ and $10^{-5}$ for another $80k$ and $60k$ iterations.}, and train the model with $10^{-3}$ learning rate for the first $280k$ iterations, then $10^{-4}$ and $10^{-5}$ for another $80k$ and $40k$ iterations, respectively.

Table \ref{tab:coco} shows the results on MS COCO {\tt test-dev} set. RefineDet320 with VGG-16 produces $29.4\%$ AP that is better than all other methods based on VGG-16 (\eg, SSD512$^*$ \cite{DBLP:conf/eccv/LiuAESRFB16} and OHEM++ \cite{DBLP:conf/cvpr/ShrivastavaGG16}). The accuracy of RefineDet can be improved to $33.0\%$ by using larger input size (\ie, $512\times512$), which is much better than several modern object detectors, \eg, Faster R-CNN \cite{DBLP:journals/pami/RenHG017} and SSD512$^*$ \cite{DBLP:conf/eccv/LiuAESRFB16}. Meanwhile, using ResNet-101 can further improve the performance of RefineDet, \ie, RefineDet320 with ResNet-101 achieves $32.0\%$ AP and RefineDet512 achieves $36.4\%$ AP, exceeding most of the detection methods except Faster R-CNN w TDM \cite{DBLP:journals/corr/ShrivastavaSMG16}, Deformable R-FCN \cite{DBLP:conf/iccv/DaiQXLZHW17}, RetinaNet800 \cite{DBLP:conf/iccv/LinPRK17}, umd\_det \cite{DBLP:conf/iccv/BodlaSCD17}, and G-RMI \cite{DBLP:conf/cvpr/HuangRSZKFFWSG016}. All these methods use a much bigger input images for both training and testing (\ie, $1000\times600$ or $800\times800$) than our RefineDet (\ie, $320\times320$ and $512\times512$). Similar to PASCAL VOC, we also report the multi-scale testing AP results of RefineDet for fair comparison in Table \ref{tab:coco}, \ie, $35.2\%$ (RefineDet320+ with VGG-16), $37.6\%$ (RefineDet512+ with VGG-16), $38.6\%$ (RefineDet320+ with ResNet-101) and $41.8\%$ (RefineDet512+ with ResNet-101). The best performance of RefineDet is $41.8\%$, which is the state-of-the-art, surpassing all published two-stage and one-stage approaches. Although the second best detector G-RMI \cite{DBLP:conf/cvpr/HuangRSZKFFWSG016} ensembles five Faster R-CNN models, it still produces $0.2\%$ lower AP than RefineDet using a single model. Comparing to the third and fourth best detectors, \ie, umd\_det \cite{DBLP:conf/iccv/BodlaSCD17} and RetinaNet800 \cite{DBLP:conf/iccv/LinPRK17}, RefineDet produces $1.0\%$ and $2.7\%$ higher APs. In addition, the main contribution: focal loss in RetinaNet800, is complementary to our method. We believe that it can be used in RefineNet to further improve the performance.

\subsection{From MS COCO to PASCAL VOC}
We study how the MS COCO dataset help the detection accuracy on PASCAL VOC. Since the object classes in PASCAL VOC are the subset of MS COCO, we directly fine-tune the detection models pretrained on MS COCO via subsampling the parameters, which achieves $84.0\%$ mAP (RefineDet320) and $85.2\%$ mAP (RefineDet512) on VOC 2007 {\tt test} set, and $82.7\%$ mAP (RefineDet320) and $85.0\%$ mAP (RefineDet512) on VOC 2012 {\tt test} set, shown in Table \ref{tab:coco-to-voc}. After using the multi-scale testing, the detection accuracy are promoted to $85.6\%$, $85.8\%$, $86.0\%$ and $86.8\%$, respectively. As shown in Table \ref{tab:coco-to-voc}, using the training data in MS COCO and PASCAL VOC, our RefineDet obtains the top mAP scores on both VOC 2007 and VOC 2012. Most important, our single model RefineNet512+ based on VGG-16 ranks as the top 5 on the VOC 2012 Leaderboard (see \cite{voc2012-leaderboard}), which is the best accuracy among all one-stage methods. Other two-stage methods achieving better results are based on much deeper networks (\eg, ResNet-101 \cite{DBLP:conf/cvpr/HeZRS16} and ResNeXt-101 \cite{DBLP:journals/corr/XieGDTH16}) or using ensemble mechanism.

\section{Conclusions}
In this paper, we present a single-shot refinement neural network based detector, which consists of two inter-connected modules, \ie, the ARM and the ODM. The ARM aims to filter out the negative anchors to reduce search space for the classifier and also coarsely adjust the locations and sizes of anchors to provide better initialization for the subsequent regressor, while the ODM takes the refined anchors as the input from the former ARM to regress the accurate object locations and sizes and predict the corresponding multi-class labels. The whole network is trained in an end-to-end fashion with the multi-task loss. We carry out several experiments on PASCAL VOC 2007, PASCAL VOC 2012, and MS COCO datasets to demonstrate that RefineDet achieves the state-of-the-art detection accuracy with high efficiency. In the future, we plan to employ RefineDet to detect some other specific kinds of objects, \eg, pedestrian, vehicle, and face, and introduce the attention mechanism in RefineDet to further improve the performance.

{\small
\bibliographystyle{ieee}
\bibliography{reference}
}

\newpage

%%%%%%%%%%%%%%%%%%%%%%%%%%%%%%%%%%%%%%%%%%%%%%%%%%%%%%%%%%%%%%%%%%%%%%%%%%%%%%%%%%%%%%%%
%% Supplementary Materials

\section{Complete Object Detection Results}
We show the complete object detection results of the proposed RefineDet method on the PASCAL VOC 2007 {\tt test} set, PASCAL VOC 2012 {\tt test} set and MS COCO {\tt test-dev} set in Table \ref{tab:pascal-voc-2007}, Table \ref{tab:pascal-voc-2012} and Table \ref{tab:coco}, respectively. Among the results of all published methods, our RefineDet achieves the best performance on these three detection datasets, \ie, $85.8\%$ mAP on the PASCAL VOC 2007 {\tt test} set, $86.8\%$ mAP on the PASCAL VOC 2012 {\tt test} set and $41.8\%$ AP on the MS COCO {\tt test-dev} set.

\begin{table*}
\centering
\caption{Object detection results on the PASCAL VOC 2007 {\tt test} set. All models use VGG-16 as the backbone network.}
\vspace{-2mm}
\footnotesize \setlength{\tabcolsep}{1.5pt}
\begin{tabular}{c|c|c|cccccccccccccccccccc}
\toprule[1.5pt]
Method &Data &mAP &aero &bike &bird &boat &bottle &bus &car &cat &chair &cow &table &dog &horse &mbike &person &plant &sheep &sofa &train &tv \\
\hline
RefineDet320 &07+12 &80.0 &83.9 &85.4 &81.4 &75.5 &60.2 &86.4 &88.1 &89.1 &62.7 &83.9 &77.0 &85.4 &87.1 &86.7 &82.6 &55.3 &82.7 &78.5 &88.1 &79.4\\
RefineDet512 &07+12 &81.8 &88.7 &87.0 &83.2 &76.5 &68.0 &88.5 &88.7 &89.2 &66.5 &87.9 &75.0 &86.8 &89.2 &87.8 &84.7 &56.2 &83.2 &78.7 &88.1 &82.3 \\
RefineDet320+ &07+12 &83.1 &89.5 &87.9 &84.9 &79.7 &70.0 &87.5 &89.1 &89.8 &69.8 &87.1 &76.4 &86.6 &88.6 &88.4 &85.3 &62.4 &83.7 &82.3 &89.0 &83.1 \\
RefineDet512+ &07+12 &83.8 &88.5 &89.1 &85.5 &79.8 &72.4 &89.5 &89.5 &89.9 &69.9 &88.9 &75.9 &87.4 &89.6 &89.0 &86.2 &63.9 &86.2 &81.0 &88.6 &84.4 \\
\hline
RefineDet320 &COCO+07+12 &84.0 &88.9 &88.4 &86.2 &81.5 &71.7 &88.4 &89.4 &89.0 &71.0 &87.0 &80.1 &88.5 &90.2 &88.4 &86.7 &61.2 &85.2 &83.8 &89.1 &85.5 \\
RefineDet512 &COCO+07+12 &85.2 &90.0 &89.2 &87.9 &83.1 &78.5 &90.0 &89.9 &89.7 &74.7 &89.8 &79.5 &88.7 &89.9 &89.2 &87.8 &63.1 &86.4 &82.3 &89.5 &84.7 \\
RefineDet320+ &COCO+07+12 &85.6 &90.2 &89.0 &87.6 &84.6 &78.0 &89.4 &89.7 &89.9 &74.7 &89.8 &80.5 &89.0 &89.7 &89.6 &87.8 &65.5 &87.9 &84.2 &88.6 &86.3 \\
RefineDet512+ &COCO+07+12 &85.8 &90.4 &89.6 &88.2 &84.9 &78.3 &89.8 &89.9 &90.0 &75.9 &90.0 &80.0 &89.8 &90.3 &89.6 &88.3 &66.2 &87.8 &83.5 &89.3 &85.2 \\
\bottomrule[1.5pt]
\end{tabular}
\label{tab:pascal-voc-2007}
\end{table*}

\begin{table*}
\centering
\caption{Object detection results on the PASCAL VOC {\tt 2012 test} set. All models use VGG-16 as the backbone network.}
\vspace{-2mm}
\footnotesize \setlength{\tabcolsep}{1.2pt}
\begin{tabular}{c|c|c|cccccccccccccccccccc}
\toprule[1.5pt]
Method &Data &mAP &aero &bike &bird &boat &bottle &bus &car &cat &chair &cow &table &dog &horse &mbike &person &plant &sheep &sofa &train &tv \\
\hline
RefineDet320 &07++12 &78.1 &90.4 &84.1 &79.8 &66.8 &56.1 &83.1 &82.7 &90.7 &61.7 &82.4 &63.8 &89.4 &86.9 &85.9 &85.7 &53.3 &84.3 &73.1 &87.4 &73.9 \\
RefineDet512 &07++12 &80.1 &90.2 &86.8 &81.8 &68.0 &65.6 &84.9 &85.0 &92.2 &62.0 &84.4 &64.9 &90.6 &88.3 &87.2 &87.8 &58.0 &86.3 &72.5 &88.7 &76.6 \\
RefineDet320+ &07++12 &82.7 &92.0 &88.4 &84.9 &74.0 &69.5 &86.0 &88.0 &93.3 &67.0 &86.2 &68.3 &92.1 &89.7 &88.9 &89.4 &62.0 &88.5 &75.9 &90.0 &80.0 \\
RefineDet512+ &07++12 &83.5 &92.2 &89.4 &85.0 &74.1 &70.8 &87.0 &88.7 &94.0 &68.6 &87.1 &68.2 &92.5 &90.8 &89.4 &90.2 &64.1 &89.8 &75.2 &90.7 &81.1 \\
\hline
RefineDet320 &COCO+07++12 &82.7 &93.1 &88.2 &83.6 &74.4 &65.1 &87.1 &87.1 &93.7 &67.4 &86.1 &69.4 &91.5 &90.6 &91.4 &89.4 &59.6 &87.9 &78.1 &91.1 &80.0 \\
RefineDet512 &COCO+07++12 &85.0 &94.0 &90.0 &86.9 &76.9 &74.1 &89.7 &89.8 &94.2 &69.7 &90.0 &68.5 &92.6 &92.8 &91.5 &91.4 &66.0 &91.2 &75.4 &91.8 &83.0 \\
RefineDet320+ &COCO+07++12 & 86.0 &94.2 &90.2 &87.7 &80.4 &74.9 &90.0 &91.7 &94.9 &71.9 &89.8 &71.7 &93.5 &91.9 &92.4 &91.9 &66.5 &91.5 &79.1 &92.8 &83.9 \\
RefineDet512+ &COCO+07++12 &86.8 &94.7 &91.5 &88.8 &80.4 &77.6 &90.4 &92.3 &95.6 &72.5 &91.6 &69.9 &93.9 &93.5 &92.4 &92.6 &68.8 &92.4 &78.5 &93.6 &85.2 \\
\bottomrule[1.5pt]
\end{tabular}
\label{tab:pascal-voc-2012}
\end{table*}

\begin{table*}
\centering
\caption{Object detection results on the MS COCO {\tt test-dev} set.}
\vspace{-2mm}
\footnotesize \setlength{\tabcolsep}{7.0pt}
\begin{threeparttable}
\begin{tabular}{c|c|ccc|ccc|ccc|ccc}
\toprule[1.5pt]
Method &Net &AP &AP$_{50}$ &AP$_{75}$ &AP$_{\it S}$ &AP$_{\it M}$ &AP$_{\it L}$ &AR$_{1}$ &AR$_{10}$ &AR$_{100}$ &AR$_{\it S}$ &AR$_{\it M}$ &AR$_{\it L}$\\
\hline
RefineDet320 &VGG-16 &29.4 &49.2 &31.3 &10.0 &32.0 &44.4 &26.2 &42.2 &45.8 &18.7 &52.1 &66.0\\
RefineDet512 &VGG-16 &33.0 &54.5 &35.5 &16.3 &36.3 &44.3 &28.3 &46.4 &50.6 &29.3 &55.5 &66.0 \\
RefineDet320 &ResNet-101 &32.0 &51.4 &34.2 &10.5 &34.7 &50.4 &28.0 &44.0 &47.6 &20.2 &53.0 &69.8 \\
RefineDet512 &ResNet-101 &36.4 &57.5 &39.5 &16.6 &39.9 &51.4 &30.6 &49.0 &53.0 &30.0 &58.2 &70.3 \\
RefineDet320+ &VGG-16 &35.2 &56.1 &37.7 &19.5 &37.2 &47.0 &30.1 &49.6 &57.4 &36.2 &62.3 &72.6 \\
RefineDet512+ &VGG-16 &37.6 &58.7 &40.8 &22.7 &40.3 &48.3 &31.4 &52.4 &61.3 &41.6 &65.8 &75.4\\
RefineDet320+ &ResNet-101 &38.6 &59.9 &41.7 &21.1 &41.7 &52.3 &32.2 &52.9 &61.1 &40.2 &66.2 &77.1 \\
RefineDet512+ &ResNet-101 &41.8 &62.9 &45.7 &25.6 &45.1 &54.1 &34.0 &56.3 &65.5 &46.2 &70.2 &79.8 \\
\bottomrule[1.5pt]
\end{tabular}
\end{threeparttable}
\label{tab:coco}
\end{table*}

\section{Qualitative Results}
\begin{figure*}
\centering
\includegraphics[width=0.98\textwidth]{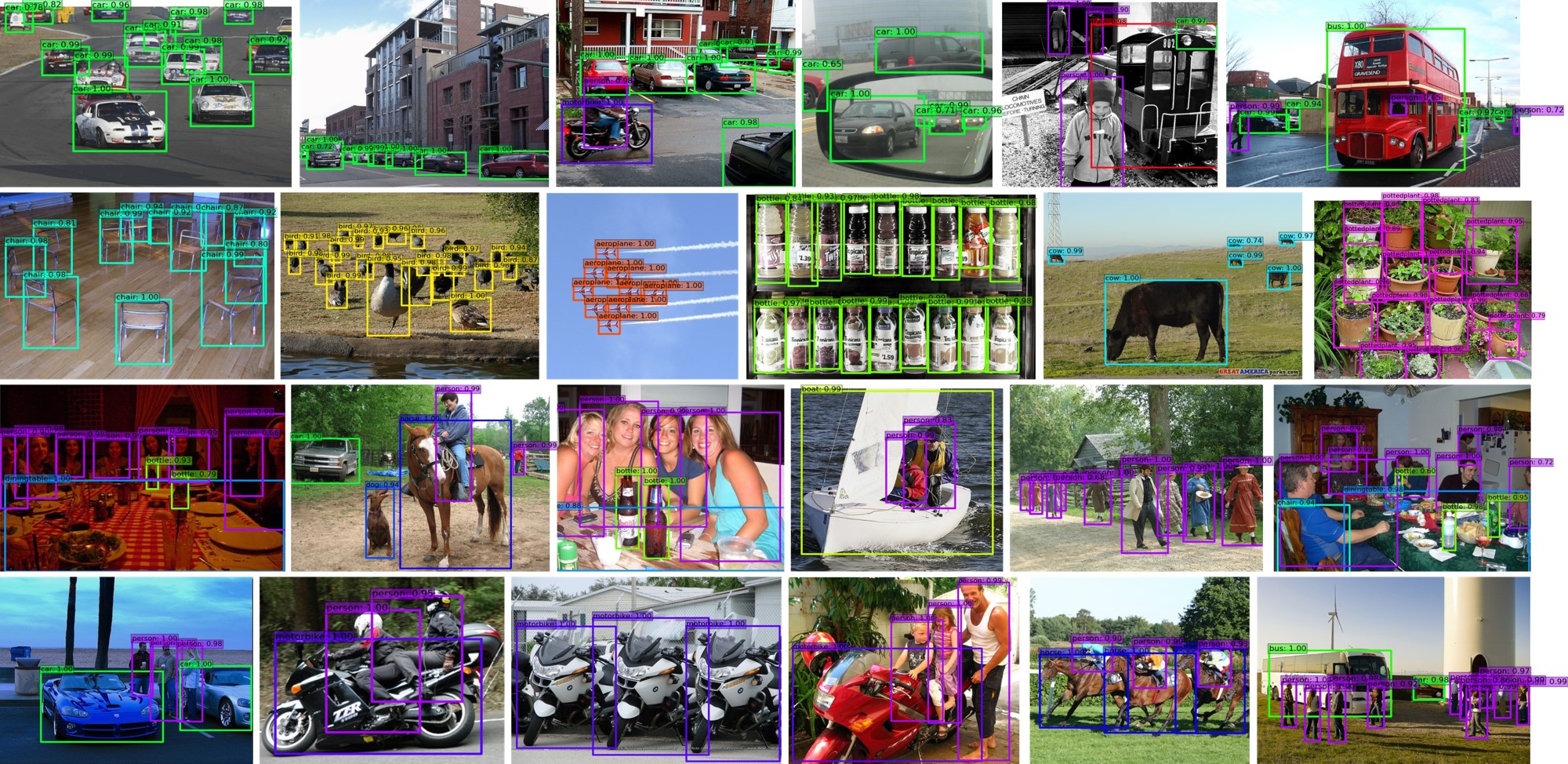}
\caption{Qualitative results of RefineDet512 on the PASCAL VOC 2007 {\tt test} set (corresponding to $85.2\%$ mAP). VGG-16 is used as the backbone network. The training data is 07+12+COCO.}
\label{fig:pascal-voc-2007}
\end{figure*}

\begin{figure*}
\centering
\includegraphics[width=0.98\textwidth]{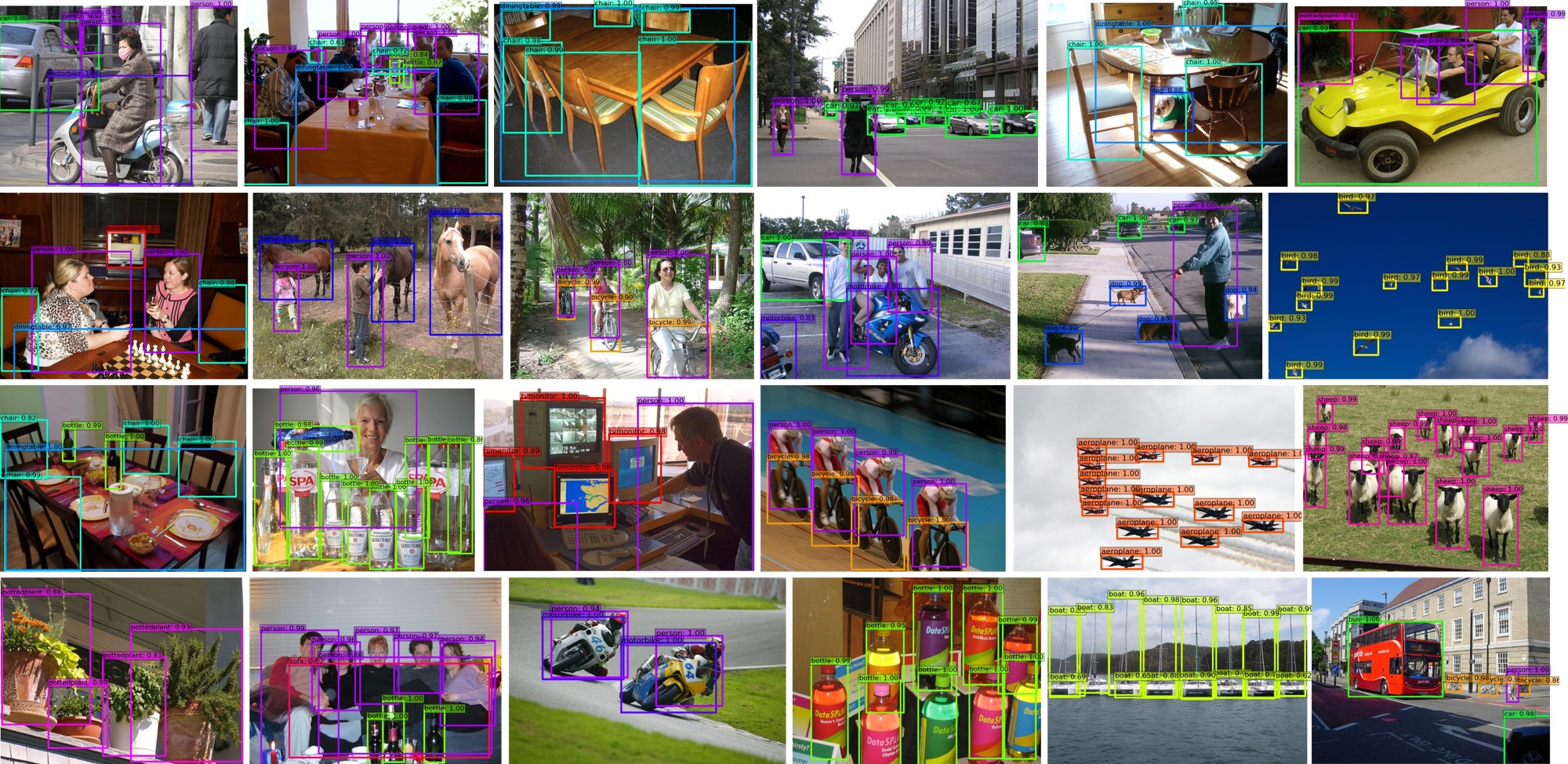}
\caption{Qualitative results of RefineDet512 on the PASCAL VOC 2012 {\tt test} set (corresponding to $85.0\%$ mAP). VGG-16 is used as the backbone network. The training data is 07++12+COCO.}
\label{fig:pascal-voc-2012}
\end{figure*}

\begin{figure*}
\centering
\includegraphics[width=0.98\textwidth]{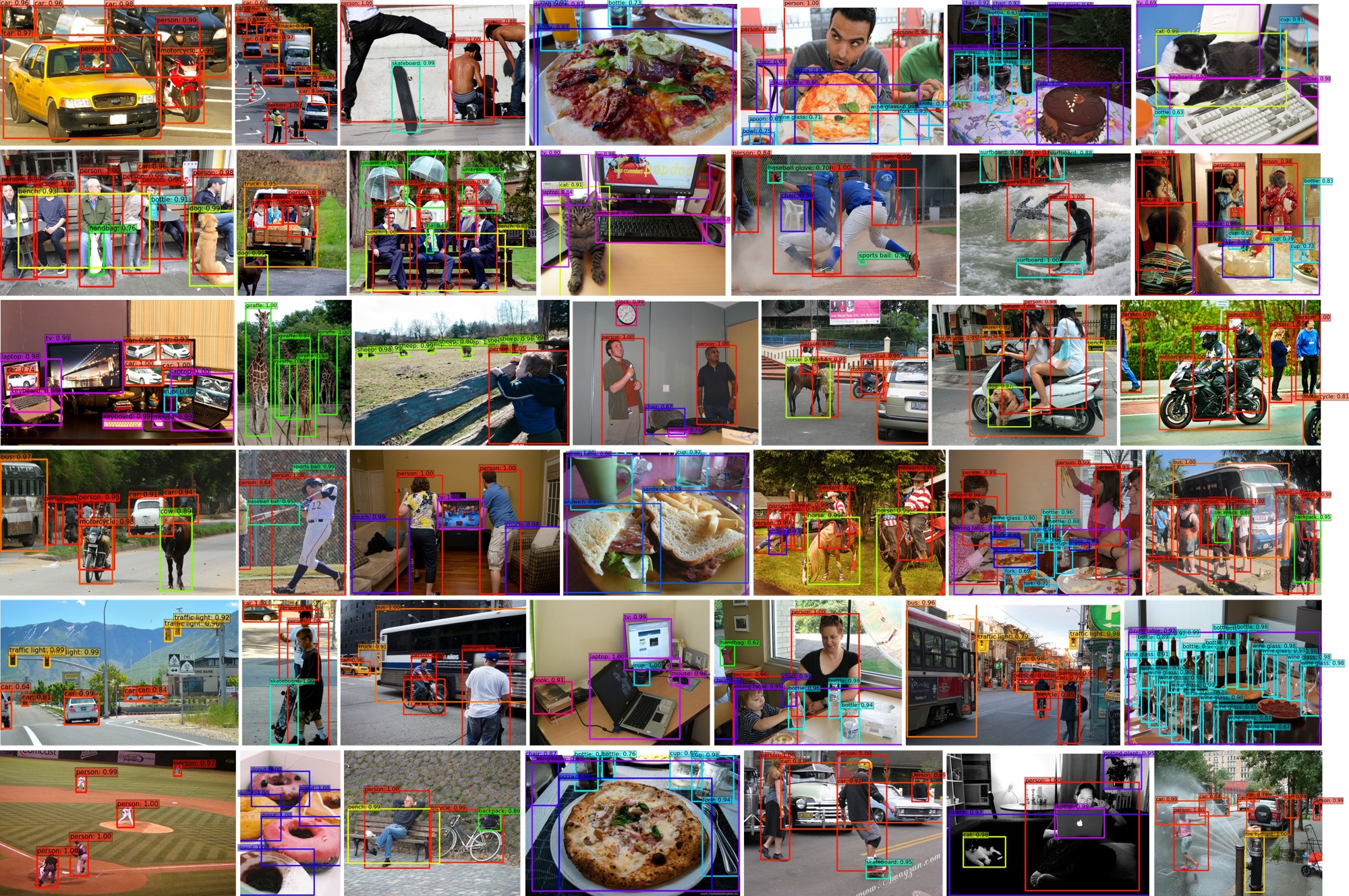}
\caption{Qualitative results of RefineDet512 on the MS COCO {\tt test-dev} set (corresponding to $36.4\%$ mAP). ResNet-101 is used as the backbone network. The training data is COCO {\tt trainval35k}.}
\label{fig:coco}
\end{figure*}

We show some qualitative results on the PASCAL VOC 2007 {\tt test} set, the PASCAL VOC 2012 {\tt test} set and the MS COCO {\tt test-dev} in Figure \ref{fig:pascal-voc-2007}, Figure \ref{fig:pascal-voc-2012}, and Figure \ref{fig:coco}, respectively. We only display the detected bounding boxes with the score larger than $0.6$. Different colors of the bounding boxes indicate different object categories. Our method works well with the occlusions, truncations, inter-class interference and clustered background.

%------------------------------------------------------------------------
\section{Detection Analysis on PASCAL VOC 2007}
\begin{figure*}
\centering
\includegraphics[width=0.96\textwidth]{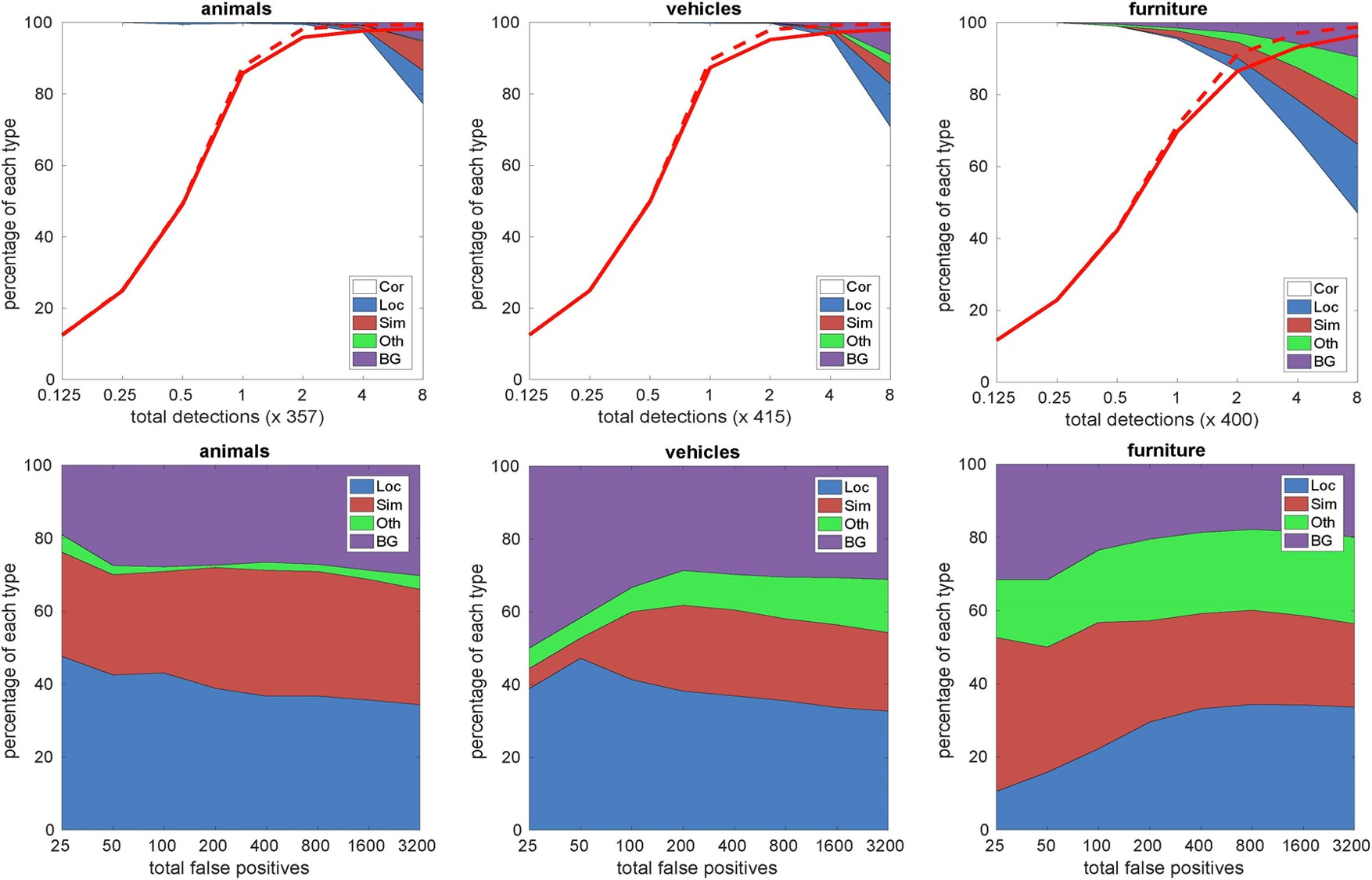}
\vspace{-2mm}
\caption{Visualization of the performance of RefineDet512 on animals, vehicles, and furniture classes in the VOC 2007 {\tt test} set. The top row shows the cumulative fraction of detections that are correct (Cor) or false positive due to poor localization (Loc), confusion with similar categories (Sim), with others (Oth), or with background (BG). The solid red line reflects the change of recall with strong criteria ($0.5$ jaccard overlap) as the number of detections increases. The dashed red line is using the ``weak'' criteria ($0.1$ jaccard overlap). The bottom row shows the distribution of the top-ranked false positive types.}
\label{fig:error}
\end{figure*}

We use the detection analysis tool\footnote{\url{http://web.engr.illinois.edu/~dhoiem/projects/detectionAnalysis/}} to understand the performance of two RefineDet models (\ie, RefineDet320 and RefineDet512) clearly. Figure \ref{fig:error} shows that RefineDet can detect various object categories with high quality (large white area). The majority of its confident detections are correct. The recall is around $95\%$-$98\%$, and is much higher with ``weak'' ($0.1$ jaccard overlap) criteria. Compared to SSD, RefineDet reduces the false positive errors at all aspects: (1) RefineDet has less localization error (Loc), indicating that RefineDet can localize objects better because it uses two-step cascade to regress the objects. (2) RefineDet has less confusion with background (BG), due to the negative anchor filtering mechanism in the anchor refinement module (ARM). (3) RefineDet has less confusion with similar categories (Sim), benefiting from using two-stage features to describe the objects, \ie, the features in the ARM focus on the binary classification (being an object or not), while the features in the object detection module (ODM) focus on the multi-class classification (background or object classes).

\begin{figure*}
\centering
\includegraphics[width=0.99\textwidth]{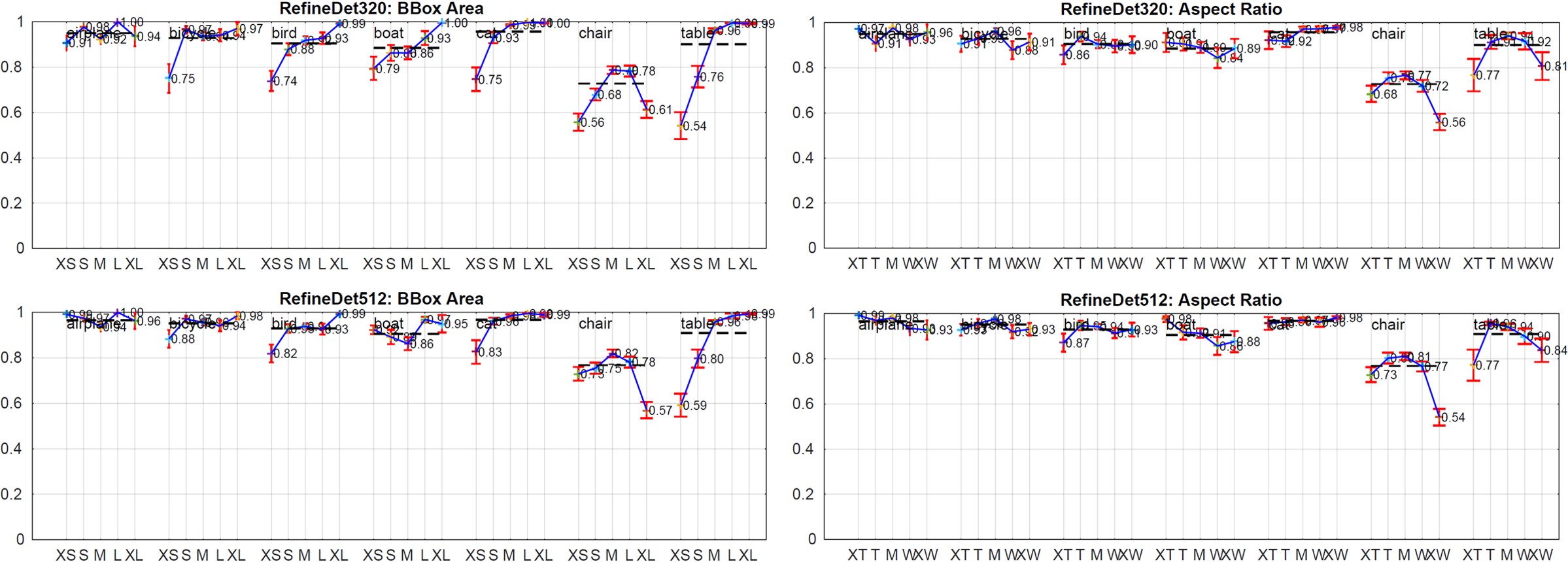}
\caption{Sensitivity and impact of different object characteristics on the VOC 2007 {\tt test} set. The plot on the left shows the effects of BBox Area per category, and the right plot shows the effect of Aspect Ratio. Key: BBox Area: XS=extra-small; S=small; M=medium; L=large; XL =extra-large. Aspect Ratio: XT=extra-tall/narrow; T=tall; M=medium; W=wide; XW =extra-wide.}
\label{fig:characteristics}
\end{figure*}

Figure \ref{fig:characteristics} demonstrates that RefineDet is robust to different object sizes and aspect ratios. This is not surprising because the object bounding boxes are obtained by the two-step cascade regression, \ie, the ARM diversifies the default scales and aspect ratios of anchor boxes so that the ODM is able to regress tougher objects (\eg, extra-small, extra-large, extra-wide and extra-tall). However, as shown in Figure \ref{fig:characteristics}, there is still much room to improve the performance of RefineDet for small objects, especially for the chairs and tables. Increasing the input size (\eg, from $320\times320$ to $512\times512$) can improve the performance for small objects , but it is only a temporary solution. Large input will be a burden on running speed in inference. Therefore, detecting small objects is still a challenge task and needs to be further studied.

\end{document}